\documentclass{article}
\usepackage{amsmath,epsfig,graphicx,multirow,float, epstopdf}
\usepackage[preprint]{spconfa4}

\begin{document}\sloppy

\def\x{{\mathbf x}}
\def\L{{\cal L}}

\title{Matrix Smoothing: A Regularization for DNN with Transition Matrix \\under Noisy Labels}

\name{Xianbin Lv$^{1, 2, \ast}$, Dongxian Wu$^{1, 2, \ast}$, Shu-Tao Xia$^{1, 2,
\ddagger}$
\thanks{$^\ast$Equal contribution.}
\thanks{$^{\ddagger}$Corresponding author: Shu-Tao Xia}
\thanks{
This work is supported in part by the National Key Research and Development Program of China under Grant 2018YFB1800204, the National Natural Science Foundation of China under Grant 61771273, the R\&D Program of Shenzhen under Grant JCYJ20180508152204044, and the research fund of PCL Future Regional Network Facilities for Large-scale Experiments and Applications (PCL2018KP001).}}

\address{$^1$Tsinghua Shenzhen International Graduate School, Tsinghua University, Shenzhen, China\\
$^2$PCL Research Center of Networks and Communications, Peng Cheng Laboratory, Shenzhen, China
\\
lv.xianbin@foxmail.com; wu-dx16@mails.tsinghua.edu.cn; xiast@sz.tsinghua.edu.cn}

\maketitle

\begin{abstract}
Training deep neural networks (DNNs) in the presence of noisy labels is an important and challenging task. Probabilistic modeling, which consists of a classifier and a transition matrix, depicts the transformation from true labels to noisy labels and is a promising approach. However, recent probabilistic methods directly apply transition matrix to DNN, neglect DNN's susceptibility to overfitting, and achieve unsatisfactory performance, especially under the uniform noise. In this paper, inspired by label smoothing, we proposed a novel method, in which a smoothed transition matrix is used for updating DNN, to restrict the overfitting of DNN in probabilistic modeling. Our method is termed \textit{Matrix Smoothing}. We also empirically demonstrate that our method not only improves the robustness of probabilistic modeling significantly, but also even obtains a better estimation of the transition matrix.
\end{abstract}
\section{Introduction}
\label{sec:intro}
Deep neural networks (DNNs) have achieved great success on many tasks, \textit{e.g.}  speech recognition \cite{hinton2012deep}, computer vision \cite{krizhevsky2012imagenet,he2016deep} and natural language processing \cite{vaswani2017attention}. To obtain the satisfactory performance, DNNs in supervised learning rely on massive training samples with high-quality annotations such as ImageNet \cite{deng2009imagenet}. However, labeling large-scale datasets is a costly and error-prone procedure. Therefore, training DNNs in the presence of noisy labels has become a task of great importance in practice.

\begin{figure}[t]
    \centering
    \includegraphics[width=0.8\linewidth]{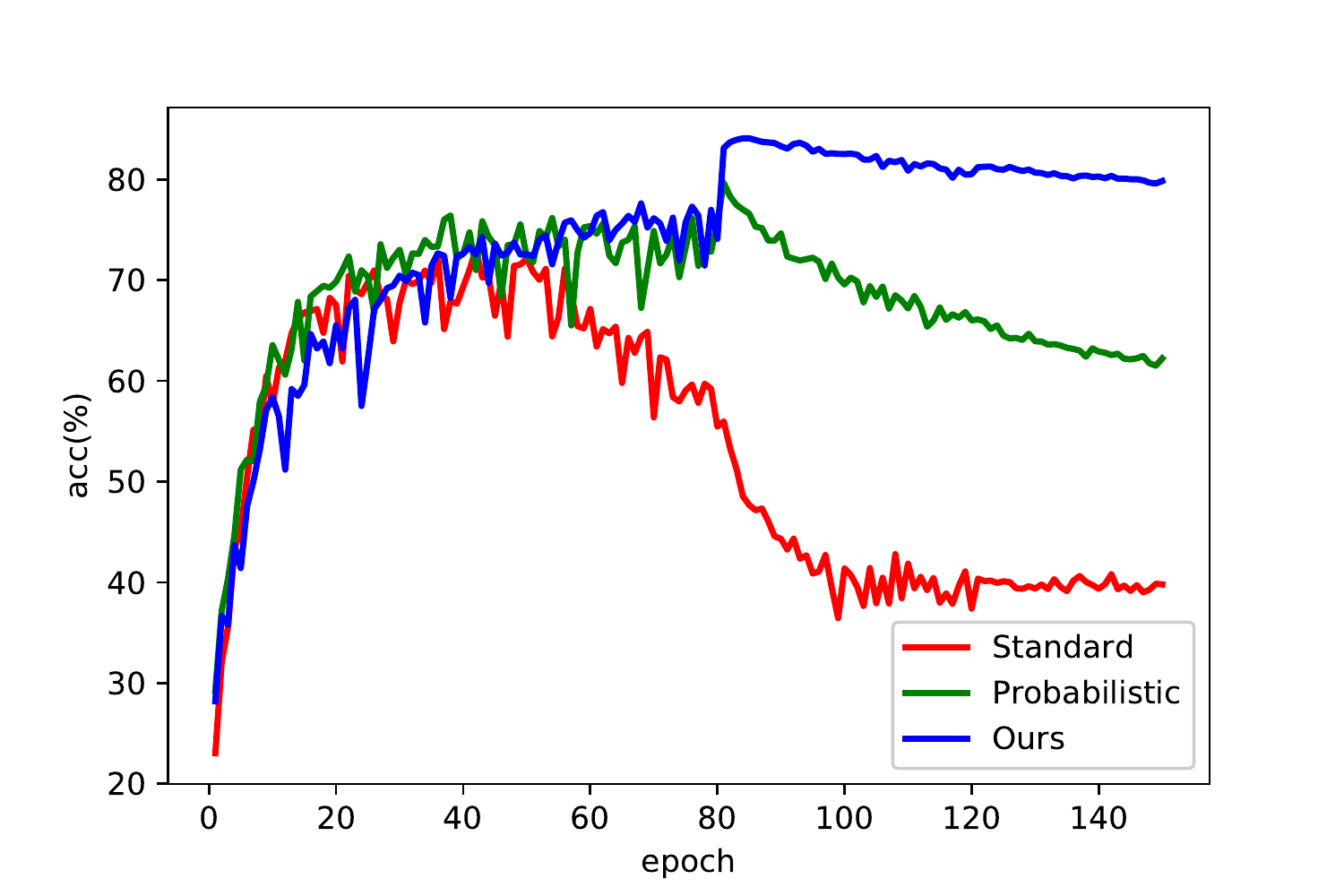}
    \caption{The accuracy(\%) on the clean test set. The standard DNN memorizes noisy labels and suffers from severe overfitting (red curve), while the probabilistic modeling with a transition matrix makes DNN achieve better performance (green curve) but still suffers from overfitting, especially after learning rate decays (80$th$ epoch). Our method has the best performance (blue curve).}
    \label{fig:learning_curve}
    \vspace{-0.5cm}
\end{figure}

Recently, several studies \cite{zhang2016understanding,ma2018dimensionality,arpit2017closer} focused on the dynamics of DNN learning with noisy labels. Zhang \textit{et al.} \cite{zhang2016understanding} argued that DNNs first memorize the training data with clean labels and subsequently memorize data with noisy labels, which leads to severe overfitting. To alleviate this problem, numerous approaches \cite{tanaka2018joint,wang2018iterative,ma2018dimensionality,wang2019symmetric} have been proposed to train DNNs with noisy labels.
Since depicting the transformation from true labels to noisy labels, probabilistic modeling is a promising method in robust training \cite{sukhbaatar2014training,patrini2017making,goldberger2016training} and crowdsourcing \cite{tanno2019learning}.
The typical probabilistic modeling consists of a classifier,  which predicts a probability distribution over a set of classes, and a transition matrix, which is usually unknown and requires estimation.
Patrini \textit{et al.} \cite{patrini2017making} proposed to estimate the transition matrix through a DNN which is trained under label noise. Bekker and Goldberger \cite{bekker2016training} derived a learning scheme based on the EM algorithm, which estimated the matrix in every M-step. Goldberger and Ben-Reuven \cite{goldberger2016training} introduced another softmax layer to the network, so as to learn the matrix and the model simultaneously. Sukhbaatar \textit{et al.} \cite{sukhbaatar2014training} emphasized the necessity to add a regularization to the ill-posed matrix estimation. In conclusion, recently published studies only focus on the estimation of the unknown transition matrix, and just regard DNN as the classifier in modeling.

However, DNNs are prone to overfit noisy labels, and probabilistic modeling only resolves this problem partially. For example, even if the accurate transition matrix is given, DNN still suffers from severe overfitting as shown in Figure \ref{fig:learning_curve}(green curve), especially after learning rate decays. Unlike the previous research, this paper pays attention to a better method to update parameters of DNN in probabilistic modeling.
Inspired by label smoothing \cite{szegedy2016rethinking} in standard DNN training, we propose a novel technique called \textit{Matrix Smoothing}. It substitutes the original transition matrix with a smoothed one for updating DNN, and keeps the matrix estimation method unchanged. Further, we provide a label-correction perspective to understand the mechanism of matrix smoothing. Finally, comprehensive experiments are conducted to demonstrate the efficiency of our method in different variants of probabilistic modeling.

\section{Matrix Smoothing}
\label{sec:matrix_smoothing}

\subsection{Preliminaries}
Assume we want to train a multi-class DNN $p(y \mid \boldsymbol{x}, \boldsymbol{w})$ where $\boldsymbol{x}$ is the input vector, $\boldsymbol{w}$ is the parameter-set of DNN and $y$ is the clean label that we cannot directly observe. Instead, only noisy labels are observed and denoted by $\tilde{y}$. Given the transformation from true labels to noisy labels, the probability of the noisy label $\tilde{y}$ is
\begin{equation}
p(\tilde{y} \mid \boldsymbol{x}, \boldsymbol{w}) = \sum_{i=1}^c p(\tilde{y}\mid y=i,\boldsymbol{x})p(y=i \mid \boldsymbol{x}, \boldsymbol{w}),
\end{equation}
where $c$ is the number of classes. In a common assumption, the noise is conditionally independent of inputs given the true labels so that
\begin{equation}
p(\tilde{y}=j \mid y=i, \boldsymbol{x}) = p(\tilde{y}=j \mid y=i) = T_{ij},
\end{equation}
where $T$ is the transition matrix for noise modeling. In general, this label noise is defined to be class dependent. The noise is uniform with rate $\eta$, if $T_{ij} = 1 - \eta$ for $i = j$, and $T_{ij} = \frac{\eta}{c-1}$ for $i \neq j$.

\begin{figure}[t]
    \centering
    \includegraphics[width=0.65\linewidth]{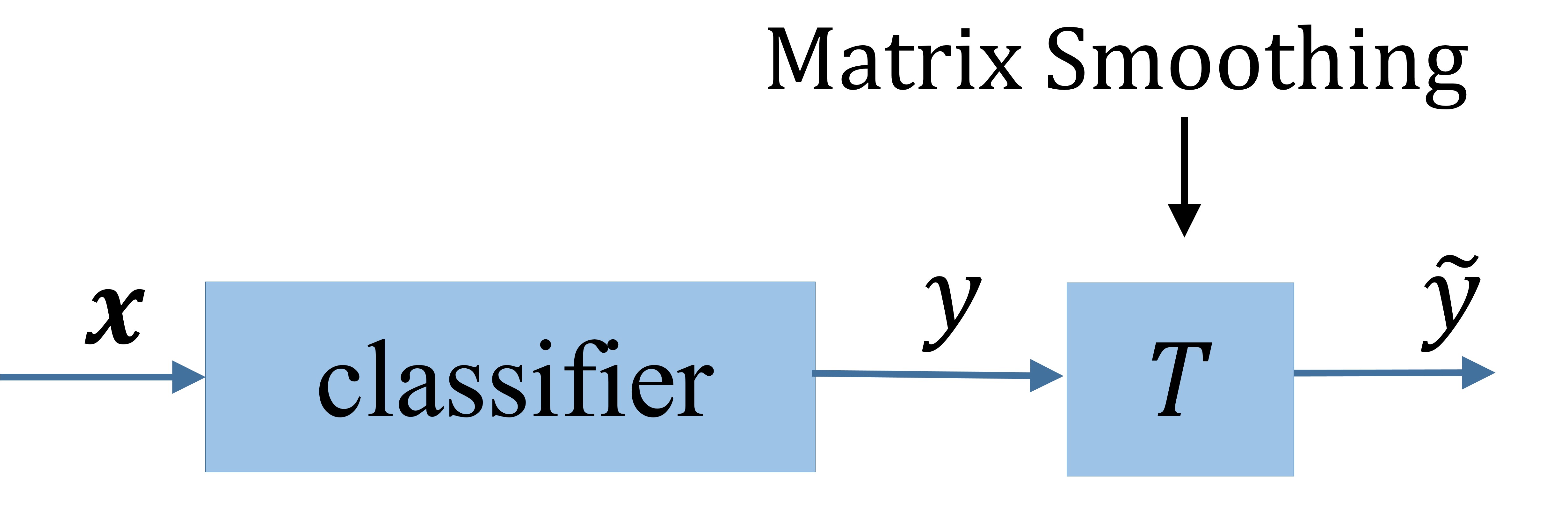}
    \caption{The diagram of the probabilistic modeling with a transition matrix, which depicts  the transformation from true labels to noisy labels. We apply matrix smoothing to the transition matrix when updating parameters of the classifier, \textit{i.e.}, DNN. }
    \label{fig:probabilistic_modeling}
    \vspace{0.1cm}
\end{figure}

In practice, we are only given the inputs $\boldsymbol{x}_1, \cdots, \boldsymbol{x}_N$ and their corresponding noisy labels $\tilde{y}_1, \cdots, \tilde{y}_n$. During training, the loss for DNN combined with transition matrix $T$ is 
\begin{equation}
\mathcal{L}_T = \frac{1}{N}\sum_{n=1}^{N}-\log \sum_{i}^{c} p(\tilde{y} \mid y=i) p(y=i\mid \boldsymbol{x}_n,\boldsymbol{w}).
\end{equation}
The model is illustrated in Figure\ref{fig:probabilistic_modeling}. After training, DNN becomes modestly robust to noisy labels as shown in Figure \ref{fig:learning_curve}. However, this method still achieves unsatisfactory performance in the clean test set, especially under training with the uniform noise.

\subsection{Label Smoothing}

Label smoothing has been widely utilized in many tasks, including image classification, language translation, and speech recognition \cite{muller2019does}. It replaces the hard targets with soft targets that are a weighted average of the hard targets and the uniform distribution over labels as illustrated in Figure \ref{fig:smoothing_method}, and improves generalization and learning speed of DNNs. Recently, Jenni demonstrated label smoothing also slightly improves robustness of DNNs to noisy labels\cite{jenni2018deep}. Unfortunately, it is incompatible with probabilistic modeling and degrades the performance on the clean test set.

\subsection{Our Method}

\begin{figure}[t]
    \centering
    \includegraphics[width=0.65\linewidth]{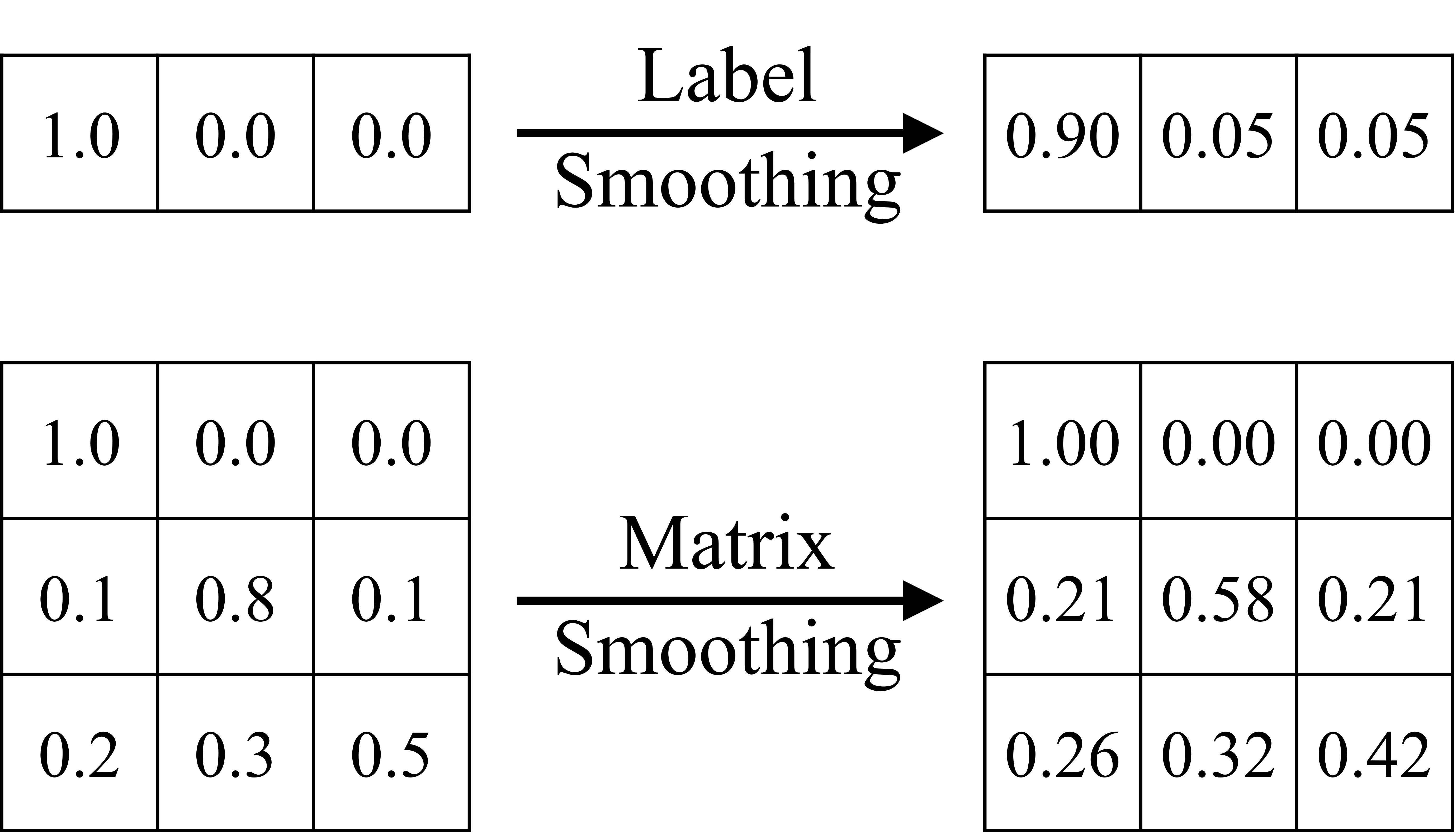}
    \caption{The diagram of label smoothing (LS, the upper one) and matrix smoothing (MS, the lower one). LS smooths the labels as new targets in supervised learning. MS smooths the transition matrix in probabilistic modeling instead.}
    \label{fig:smoothing_method}
\end{figure}

Inspired by label smoothing in standard DNN training, we introduce a novel technique to the probabilistic modeling. Assuming the true transition matrix $T$ is given, we always update parameters of DNN through $T$ directly in the conventional probabilistic modeling. Instead, we substitute the original transition matrix $T$ with a smoothed one $S$, that is,
\begin{equation}
    S_{ij} = \frac{T_{ij}^\beta}{\sum_l T_{il}^\beta},
\end{equation}
where $\beta \in (0, 1]$. Figure \ref{fig:smoothing_method} exhibits an example of matrix smoothing with a $3 \times 3$ matrix and $\beta=0.5$.

\subsubsection{The Smoothing Methods}
\label{sec:smothing methods}
Actually, there are several methods to smooth the matrix. For example, we could utilize a weighted average of the original transition matrix and a special matrix whose all elements are same, that is,

\begin{equation}
    S = \gamma T + (1 - \gamma) \frac{1}{c} \boldsymbol{1},
\end{equation}
where $\gamma \in (0, 1]$ and $\boldsymbol{1}$ is a matrix whose elements are all equal to $1$. This method is similar to label smoothing. Since it is a linear weighted average of the original matrix and a matrix of the uniform distribution, we denote it as ``the linear method''. However, if there are zero elements in the transition matrix, e.g. the first row in Figure \ref{fig:smoothing_method}, the second smoothing method introduces incorrect information and misleads DNN to believe the impossible noise. By contrast, our method does not change elements in the matrix if they are equal to zero.

Another method is to introduce a hyper-parameter $\gamma$ and the smoothed matrix $S$ becomes
\begin{equation}
    S_{ij} = \frac{\exp( b_{ij} / \gamma)}{\sum_l \exp(b_{il} / \gamma)},
\end{equation}
where $\gamma \in [1, \infty)$ and $b_{ij} = \log(T_{ij})$. This formula is related to model distilling \cite{hinton2015distilling}, and  $\gamma$ is the hyper-parameter of the temperature. And we call this ``the temperature method''. However, this method cannot deal with the situation in which any element is zero, since $\log 0 = - \infty$.

These methods have the same effect for training DNN with a suitable selection of hyper-parameters under the uniform noise. We will compare these three smoothing methods under the asymmetric noise in Section \ref{sec:compare_with_othermethods}.
\subsubsection{Combination with Matrix Estimation}

However, the transition matrix is not always accessible in practice. An accurate estimation of the transition matrix is of significance in realistic scenarios. There are several proposed estimation methods including direct estimation and joint learning with DNN. Fortunately, it is easy to combine matrix smoothing with these matrix estimation methods.

Patrini \textit{et al.} \cite{patrini2017making} proposed to train DNN in a standard manner with noisy labels, estimate the transition matrix from the trained DNN and train another DNN with the estimated transition matrix. Since the transition matrix is fixed in the second round of training, we could apply matrix smoothing directly. Goldberger and Ben-Reuven \cite{goldberger2016training} introduced another softmax layer to the network for the transition matrix, so as to learn the matrix and DNN simultaneously. Sukhbaatar \cite{sukhbaatar2014training} has a similar training strategy. For these joint learning strategies, we could decouple the learning steps: first, we update DNN with matrix smoothing when the matrix is fixed; second, we utilize the original method to update the matrix. In conclusion, matrix smoothing is compatible with these matrix estimation methods.

\section{Understanding Matrix Smoothing}
\subsection{A Label-Correction Perspective}
Matrix smoothing is a simple yet effective method to restrict DNN from overfitting in probabilistic modeling. Here, we provide an explanation of its mechanism. To explore the mechanism of matrix smoothing, it is convenient to rewrite the probabilistic modeling into a label-correction form. Here, we consider the loss of one sample, \textit{i.e.} $(\boldsymbol{x},\tilde{y} = j)$, and the result is
\begin{equation}
\begin{aligned}
& \frac{\partial}{\partial \boldsymbol{w}} \big( - \log \sum_{i=1}^{c} T_{ij} p(y=i \mid \boldsymbol{x}, \boldsymbol{w}) \big) \\
=& - \sum_{i=1}^c \frac{T_{ij}}{\sum_{k} T_{kj} p(y=k|\boldsymbol{x}, \boldsymbol{w})} \frac{\partial}{\partial \boldsymbol{w}}p(y=i|\boldsymbol{x}, \boldsymbol{w})\\
=& -\sum_{i=1}^c \frac{T_{ij} p(y=i|\boldsymbol{x}, \boldsymbol{w})}{\sum_{k} T_{kj}p(y=k|\boldsymbol{x}, \boldsymbol{w})} \frac{\partial}{\partial \boldsymbol{w}}\log p(y=i|\boldsymbol{x}, \boldsymbol{w})\\
=& -\sum_{i=1}^c p(y=i|\tilde{y}=j,\boldsymbol{x}, \boldsymbol{w}) \frac{\partial}{\partial \boldsymbol{w}} \log p(y=i|\boldsymbol{x}, \boldsymbol{w}),
\end{aligned}
\label{eqn:equivalent}
\end{equation}
where
\begin{equation}
\begin{aligned}
p(y=i|\tilde{y}=j,\boldsymbol{x}, \boldsymbol{w}) &= \frac{p(y=i, \tilde{y}=j \mid \boldsymbol{x}, \boldsymbol{w})}{p(\tilde{y}=j \mid \boldsymbol{x}, \boldsymbol{w})} \\
&= \frac{T_{ij} p(y=i|\boldsymbol{x}, \boldsymbol{w})}{\sum_{k} T_{kj}p(y=k|\boldsymbol{x}, \boldsymbol{w})}
\label{eqn:new_targets}
\end{aligned}
\end{equation}
is the posteriori probability for the clean label, in which a noisy label is given and the output probability of DNN is regarded as the priori probability. Training DNN in probabilistic modeling with noisy labels is equivalent to training standard DNN with modified labels as Eqn.(\ref{eqn:new_targets})

\subsection{The Mechanism of Matrix Smoothing}

For simplicity, we only consider the situation with the uniform noise\footnote{The situation with the uniform noise is more important than that with the asymmetric noise since the former is more difficult and the probabilistic modeling performs poorly with the uniform noise.}. The gradient of the loss with matrix smoothing has a similar formula as Eqn.(\ref{eqn:equivalent}), except the posteriori item, which becomes,
\begin{equation}
    q(y \mid \tilde{y},\boldsymbol{x}, \boldsymbol{w}) = \frac{\alpha p(y, \tilde{y} \mid \boldsymbol{x}, \boldsymbol{w}) + (1 - \alpha) \frac{1}{c} p(y \mid \boldsymbol{x},\boldsymbol{w})}{\alpha p(\tilde{y} \mid \boldsymbol{x}, \boldsymbol{w}) + (1 - \alpha)\frac{1}{c}},
\end{equation}
where
\begin{equation}
    \alpha = \frac{\frac{(1-\eta)^\beta}{(1-\eta)^\beta + (c-1)(\frac{\eta}{c-1})^\beta} - \frac{1}{c}}{1 - \eta - \frac{1}{c}}.
\end{equation}
When $\beta=0$, we have $\alpha = 0$, which makes modified labels become the output probability of DNN. If $\beta=1$, we have $\alpha=1$ and modified labels degenerate to the posteriori probability as Eqn.(\ref{eqn:new_targets}). When $0<\alpha<1$, modified labels are a combination of output probability and the original posterior probability. Since DNN behaves robustly to noisy labels and achieves satisfactory performance at the early stage during training as indicated in \cite{ma2018dimensionality,arpit2017closer}, it is reasonable to rely more on its outputs at the early training stage.

\section{Experiment}
\label{sec:experiment}

In this section, we comprehensively verify the effectiveness of our proposed method, Matrix Smoothing, on synthetic noisy datasets.

\subsection{Detailed Comparison of Smoothing Methods}

\subsubsection{Comparison with Label Smoothing}
\label{sec:compare_with_ls}

We first demonstrate the effectiveness of matrix smoothing, in comparison with another regularization technique, label smoothing\cite{szegedy2016rethinking}. As indicated in \cite{patrini2017making}, poor matrix estimation degrades robustness. To eliminate the influence of estimation and just study the performance limits, we assume that the transition matrix $T$ is given here.

\noindent\textbf{Datasets.}
We conduct experiments on noisy CIFAR-10. 
Noisy labels are generated by introducing the uniform noise, in which the label of a given training sample is flipped uniformly to one of the other classes with probability $\eta$. The uniform noise has been verified to be more challenging than the asymmetric noise as demonstrated in \cite{ma2018dimensionality,patrini2017making}

\noindent\textbf{Baselines.} 
Forward method \cite{patrini2017making} with a given transition matrix $T$ is regarded as the basic model. We compare the robustness of the original Forward method(FD), Forward with label smoothing\cite{szegedy2016rethinking} (FD+LS) and Forward with the proposed Matrix Smoothing (FD+MS).

\noindent\textbf{Experimental setup.}
We utilize PreActResNet18\cite{he2016identity} as the basic classifier in probabilistic modeling, which is a powerful DNN with accuracy 94.18\%(74.41\%) on the test set after trained on clean CIFAR-10(CIFAR-100). We use a batch size of $128$, a weight decay of $10^{-4}$, and SGD with the momentum of 0.9. We train the network until 150 epochs with initial learning rate of 0.1, and decrease the learning rate by dividing it by 10 after 80 and 120 epochs.

As shown in Table \ref{table:with_ls}, even though the accurate transition matrix is given, Forward still suffers severe overfitting, especially in a large noise rate. The label smoothing is incompatible with probabilistic modeling based on the transition matrix, despite the fact that it improves robustness slightly in standard DNN training with noisy labels\cite{jenni2018deep}. The proposed matrix smoothing improves the robustness of Forward remarkably by 5\% $\sim$ 20\%. This result indicates the effectiveness of our method as a regularization for training DNN with a transition matrix.

\begin{table}
\renewcommand\tabcolsep{3.0pt}
\small
\centering
\caption{Test accuracy (\%) on CIFAR-10 with varying noise rates (0.2 - 0.8). The mean accuracy ($\pm$std) over 5 repetitions of the experiments are reported, and the best results are highlighted in \textbf{bold}.} 
\label{table:with_ls}
\begin{tabular}{c|cccc}
\hline
\multirow{2}{*}{Methods} & \multicolumn{4}{c}{noise rate $\eta$} \\ \cline{2-5} 
& 0.2    & 0.4   & 0.6   & 0.8   \\ \hline
FD  & 86.23$\pm$0.64  & 75.93$\pm$1.08 & 62.13$\pm$1.65 & 41.43$\pm$1.41 \\
FD+LS  & 82.00$\pm$0.52       &69.58$\pm$0.75       & 56.16$\pm$1.72      &40.65$\pm$5.27       \\
FD+MS  & \textbf{91.21$\pm$0.40}  & \textbf{87.60$\pm$0.46} & \textbf{78.54$\pm$0.55} & \textbf{50.93$\pm$1.82} \\ \hline
\end{tabular}
\end{table}

\begin{table}
\renewcommand\tabcolsep{3.0pt}
\small
\centering
\caption{Test accuracy (\%) on CIFAR-10 with varying asymmetric noise rates (0.1 - 0.4). The mean accuracy ($\pm$std) over 5 repetitions of the experiments are reported, and the best results are highlighted in \textbf{bold}.} 
\label{table:CIFAR10-asym}
\begin{tabular}{c|cccc}
\hline
\multirow{2}{*}{Methods} & \multicolumn{4}{c}{noise rate $\eta$} \\ \cline{2-5} 
& 0.1    & 0.2   & 0.3   & 0.4\\ \hline
FD  & 92.31$\pm$0.09  & 90.05$\pm$0.11 & 88.06$\pm$0.30 & 85.73$\pm$0.26 \\
FD+L  & 92.48$\pm$0.43      &90.58$\pm$0.37 & 88.03$\pm$0.26& 86.32$\pm$0.22 \\
FD+T   & 92.11$\pm$0.12      &89.95$\pm$0.12 & 87.18$\pm$0.30 & 84.87$\pm$0.37       \\
FD+Ours  & \textbf{93.17$\pm$0.25}  & \textbf{92.20$\pm$0.32} & \textbf{91.23$\pm$0.51} & \textbf{89.69$\pm$1.12} \\ \hline
\end{tabular}
\end{table}
\subsubsection{Comparison with Other Matrix Smoothing Methods}
\label{sec:compare_with_othermethods}
Since there are several methods to smooth matrix as shown in Section \ref{sec:matrix_smoothing}, we compare these methods under the asymmetric noise\footnote{These methods have the same effect under the uniform label noise with suitable hyper-parameters. Thus, we compare them under the asymmetric noise instead.} here.

\noindent\textbf{Datasets \& Experimental setup \& Baselines.} We compare these methods on CIFAR-10 under the asymmetric noise and use the same experimental setup (architecture, prepossessing, training setting) as Section \ref{sec:compare_with_ls}. Forward method(FD) is chosen as the baseline. For fairness, we select the best hyper-parameters for every method: for our chosen method(Ours), the parameter $\beta$ is 0.5; for ``the linear method''(L) that uses all-one matrix to smooth transition matrix, the parameter $\gamma$ is 0.8; for ``the temperature method''(T), the parameter of the temperature $\gamma$ is 1.1.

The results are demonstrated in Table \ref{table:CIFAR10-asym}. Our method achieves the best performance under all the asymmetric noise rates. Although the hyper-parameters have been selected, the other methods is still unsatisfactory, which is compatible with the discussion in Section \ref{sec:matrix_smoothing}, \textit{i.e.}, the other methods would introduce incorrect information for probabilistic modeling.

\subsection{Matrix Smoothing with An Unknown Matrix}

The transition matrix is not always accessible in practice, and the estimation of the transition matrix is important in realistic scenarios. In this section, we combine several matrix estimation methods with matrix smoothing, to demonstrate that our method not only prevents DNN from overfitting, but also improves the estimation quality. 

\noindent\textbf{Datasets \& Experimental setup.} For comprehensive comparison, we utilize CIFAR-10 and CIFAR-100 with the uniform noise. We use the same experimental setup (architecture, prepossessing, training setting) as Section \ref{sec:compare_with_ls}.

\begin{table}
\renewcommand\tabcolsep{3.0pt}
\small
\centering
\caption{Test accuracy (\%) on CIFAR-10 with varying noise rates (0.2 - 0.8). The mean accuracy ($\pm$std) over 5 repetitions of the experiments are reported, and the best results are highlighted in \textbf{bold}.} 
\label{table:cifar10}
\begin{tabular}{c|cccc}
\hline
\multirow{2}{*}{Methods} & \multicolumn{4}{c}{noise rate} \\ \cline{2-5} 
& 0.2    & 0.4   & 0.6   & 0.8   \\ \hline
CE  &80.60$\pm$0.33 &62.39$\pm$0.37 &39.79$\pm$0.80 &17.58$\pm$0.42 \\
GCE&91.43$\pm$0.18 &87.00$\pm$0.07 &69.32$\pm$0.75 &24.11$\pm$0.55 \\
FD$_{est}$ &80.92$\pm$0.87 &62.13$\pm$0.44 &40.49$\pm$1.29 &17.91$\pm$0.64\\
FD$_{est}$+MS  &91.35$\pm$0.39 &87.09$\pm$0.50 &76.30$\pm$0.31 &29.60$\pm$6.60  \\
AL &79.91$\pm$0.49 &60.81$\pm$0.64 &40.01$\pm$1.21 &17.56$\pm$0.42 \\
AL+MS &\textbf{91.45$\pm$0.26}
		& \textbf{87.24$\pm$0.37} & \textbf{78.66$\pm$0.53} & \textbf{48.24$\pm$2.37} \\ \hline \hline
FD  & 86.23$\pm$0.64  & 75.93$\pm$1.08 & 62.13$\pm$1.65 & 41.43$\pm$1.41      \\
FD+MS& 91.21$\pm$0.40 & \textbf{87.60$\pm$0.46} & 78.54$\pm$0.55 & \textbf{50.93$\pm$1.82} \\ \hline 
\end{tabular}
\end{table}

\begin{table}
\renewcommand\tabcolsep{3.0pt}
\small
\centering
\caption{Test accuracy (\%) on CIFAR-100 with varying noise rates (0.2 - 0.8). The mean accuracy ($\pm$std) over 5 repetitions of the experiments are reported, and the best results are highlighted in \textbf{bold}.} 
\label{table:cifar100}
\begin{tabular}{c|cccc}
\hline
\multirow{2}{*}{Methods} & \multicolumn{4}{c}{noise rate} \\ \cline{2-5} 
& 0.2    & 0.4   & 0.6   & 0.8   \\ \hline
CE &59.48$\pm$0.51 &42.73$\pm$0.30 &25.90$\pm$0.30 &8.11$\pm$0.38 \\
GCE&68.03$\pm$0.48 &\textbf{64.37$\pm$0.41} &\textbf{56.74$\pm$0.52} &\textbf{33.44$\pm$0.84} \\
AL &61.56$\pm$0.23 &44.62$\pm$0.95 &21.19$\pm$0.59 &5.79$\pm$0.20\\
AL+MS &\textbf{69.69$\pm$0.25} &63.91$\pm$0.56 &53.93$\pm$0.47 &18.16$\pm$0.48\\ \hline \hline
FD  & 63.94$\pm$0.66  & 52.24$\pm$0.89 & 37.52$\pm$0.76 & 19.72$\pm$0.87\\
FD+MS&\textbf{69.92$\pm$0.36} &63.67$\pm$0.47 &51.36$\pm$0.43 &27.66$\pm$1.33\\ \hline 
\end{tabular}
\end{table}

\noindent\textbf{Baselines.}
Two different estimation methods for $T$ are chosen as our baselines. (1) Estimation from a pre-trained model: In \cite{patrini2017making}, a standard DNN is trained on noisy dataset, and transition matrix is estimated from some ``perfect samples''. We denote this as FD$_{est}.$(2) Adaption Layer (AL)\cite{goldberger2016training}: The transition matrix $T$ is regarded as an extra fully connected layer and train with DNN simultaneously. We initialize $T$ using an identity matrix\footnote{For numerical stability, the actual $T$ is an identity matrix added by an extremely small uniform noise.} as shown in \cite{goldberger2016training}. In the first 15 epochs, we freeze the transition matrix and only update DNN as a warmup. Then we update the matrix and DNN simultaneously. To illustrate the superiority of Forward combined with Matrix Smoothing, Cross Entropy(CE) and Generalized Cross Entropy(GCE) Loss are included as extra baselines. We exclude FD$_{est}$ on CIFAR-100, since it has bad performance.
 
\noindent\textbf{Hyper-parameter selection.}
We select suitable $\beta$ of matrix smoothing for different methods. First, we split the original noisy training set (50k images in CIFAR-10/CIFAR-100) into a new training set(40k) and a validation set(10k). We train models in 40k training set and compare hyper-parameter in the noisy 10k validation set. As the hyper-parameter is fixed, we retrain models in the total training set(50k) from scratch. As a result, $\beta=0.5$ for FD+MS, $\beta=0.8$ for AL+MS, and $\beta=0.1$ for FD$_{est}$+MS.

\subsubsection{Robustness against Noisy Labels}

As demonstrated in Table \ref{table:cifar10}, unsatisfying estimation of $T$ degrades the performance of probabilistic modeling by 5\% $\sim$ 23\% (comparing FD$_{est}$/AL with FD), which restricts its practical application. Further, even though $T$ is known, the robustness of Forward is still limited in comparison with GCE, a state-of-art robust loss. Fortunately, matrix smoothing solves these problems effectively. FD+MS has similar or even better results than GCE in CIFAR-10, and FD$_{est}$+MS and AL+MS have similar performance to FD+MS in CIFAR-10. As shown in Table \ref{table:cifar100}, GCE has best performance on CIFAR-100 since it was proposed given the situation with the uniform noise. FD is worse than GCE significantly. Combined with MS, FD has much better performance and its results are close to GCE, which also indicates the effectiveness of MS in CIFAR-100. In conclusion, the improvements brought by matrix smoothing in probabilistic modeling is significant and consistent, no matter the matrix is known or estimated.
\begin{figure}

\begin{minipage}[a]{.48\linewidth}
  \centering
  \centerline{\includegraphics[width=4.5cm]{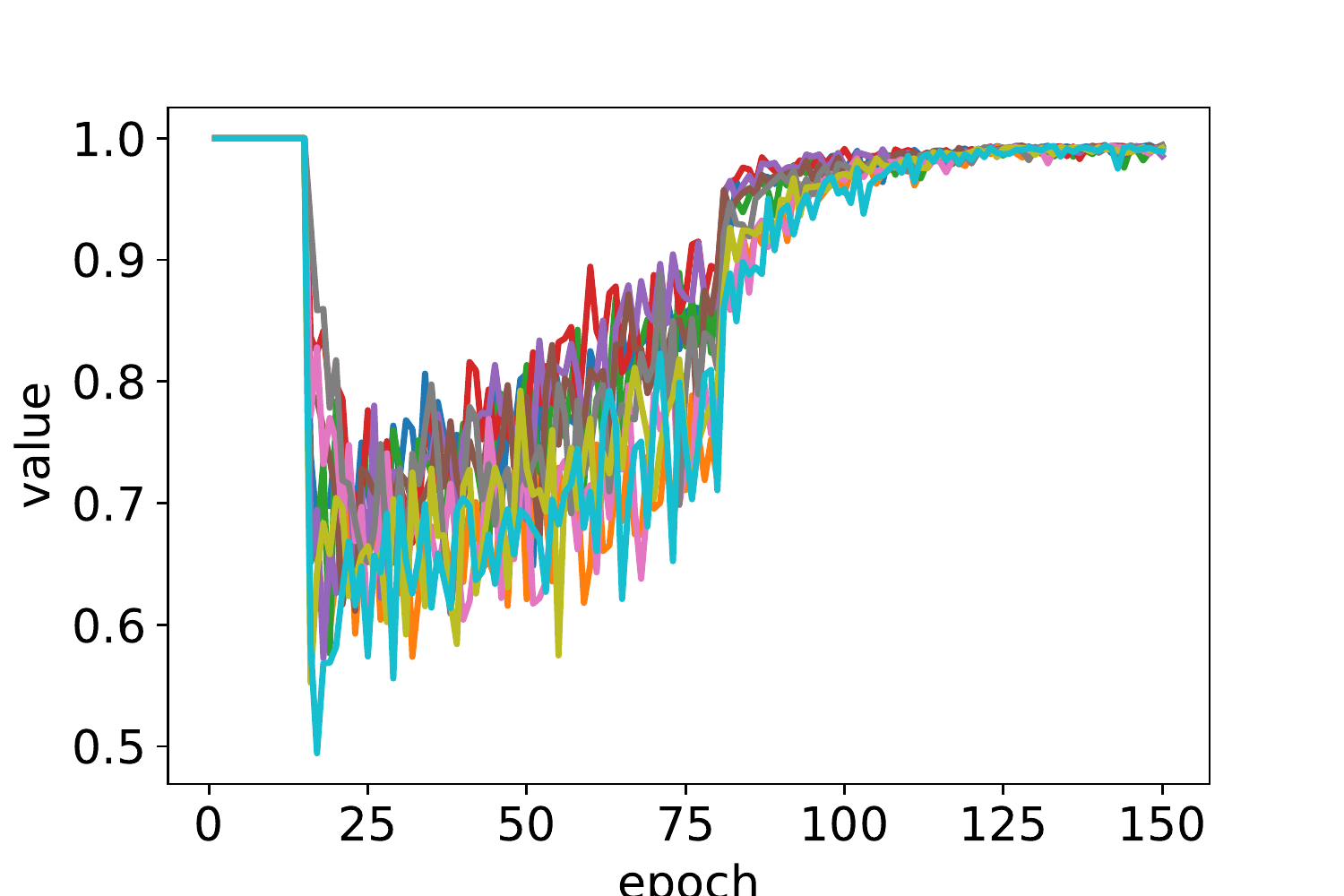}}
  \centerline{(a)}\medskip
\end{minipage}
\hfill 
\begin{minipage}[]{0.48\linewidth}
  \centering
  \centerline{\includegraphics[width=4.5cm]{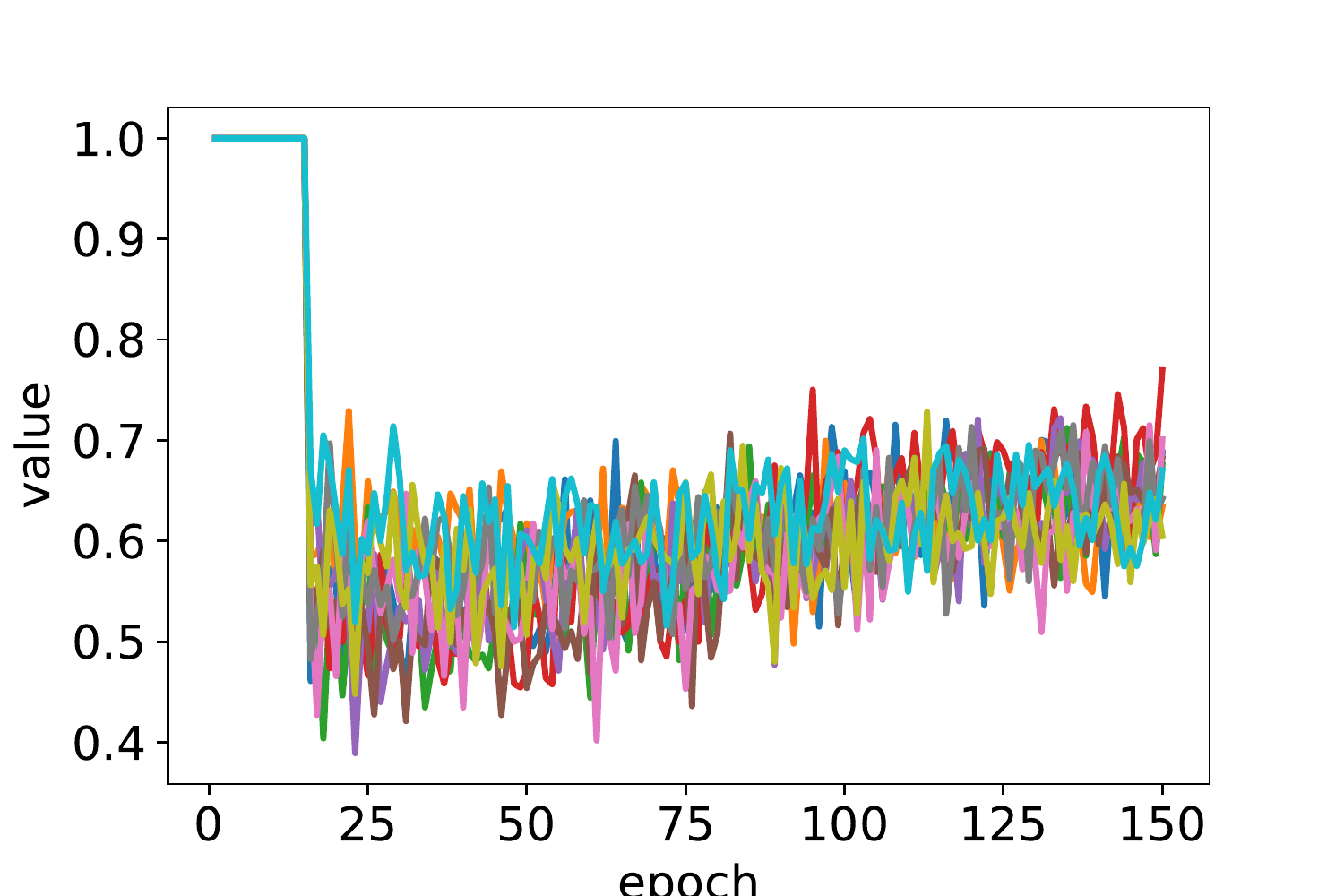}}
  \centerline{(b)}\medskip
\end{minipage}
\caption{The values of elements in the main diagonal of the transition matrix during training. The true value of these elements is 0.6. (a) indicates the result of the original AL, and (b) indicates the result of ours (AL+MS). }
\label{fig:estimateT}
\end{figure}%
\subsubsection{Estimation of Transition Matrix}

In Adaption Layer, transition matrix and DNN are updated simultaneously, and their qualities affect the updating of each other. Previously published studies only focused on the impact on DNN from matrix estimation and introduced regularization to matrix learning. Instead, our method keeps matrix learning unchanged and utilizes the smoothed matrix when updating DNN. In this section, we demonstrate the matrix estimation has satisfying performance as long as DNN is trained of high quality. 

To demonstrate the estimation quality of AL and AL+MS, we visualize the elements in the main diagonal of the learned transition matrix with 0.4 uniform noise in Figure \ref{fig:estimateT}. The true value of these elements is 0.6. For original AL, the model has unsatisfactory performance as the learned transition matrix tends to identity matrix and DNN overfits to noisy labels severely. By contrast, elements in main diagonal converge to the true value (0.6) in ours(AL+MS), which indicates that we could obtain a satisfied estimation of transition matrix just through updating DNN in a better way.

\section{Conclusion}

In this paper, we demonstrate that DNN with a transition matrix in probabilistic modeling still suffers from severe overfitting. Inspired by label smoothing, we propose a new method called \textit{Matrix Smoothing}, which replaces the original matrix with a smoothed one when updating parameters of DNN. Further, we provide a label-correction perspective for probabilistic modeling to explain the mechanism of matrix smoothing, and verify its effectiveness through comprehensive experiments. Our results in this paper remind researchers to consider not only the estimation of transition matrix in probabilistic modeling, but also the parameter updating for DNN.

\bibliographystyle{IEEEbib}
\bibliography{icme2020template}

\end{document}